\documentclass[10pt,twocolumn,letterpaper,english,table]{article}

\usepackage{pgfplots}

\usepackage{cvpr}
\usepackage{times}
\usepackage{epsfig}
\usepackage{graphicx}
\usepackage{amsmath}
\usepackage{amssymb}
\usepackage{url}

\usepackage{booktabs}       % professional-quality tables
\usepackage{algorithm}
\usepackage{algpseudocode}
\usepackage{epstopdf}
\usepackage{babel}
\usepackage{xcolor}

\usepackage{graphicx,tikz}
\usepackage{pgfplots}
\usetikzlibrary{patterns}
\usepgfplotslibrary{groupplots}
\usetikzlibrary{matrix}

\usepackage[pagebackref=true,breaklinks=true,letterpaper=true,colorlinks,bookmarks=false]{hyperref}

 \cvprfinalcopy % *** Uncomment this line for the final submission

\def\cvprPaperID{4099} % *** Enter the CVPR Paper ID here

\begin{document}

%%%%%%%%% TITLE
\title{Curls \& Whey: Boosting Black-Box Adversarial Attacks}

%% omit the following lines up until the closing ``}''.
%% Additional authors and addresses can be added with ``\and'',
%% just like the second author.
%% To save space, use either the email address or home page, not both
%\and


\author{Yucheng Shi, Siyu Wang, Yahong Han\\
College of Intelligence and Computing\\
Tianjin University, Tianjin, China\\
{\tt\small \{yucheng, syuwang, yahong\}@tju.edu.cn}}

\maketitle
\thispagestyle{empty}

%%%%%%%%% ABSTRACT
\begin{abstract}
   Image classifiers based on deep neural networks suffer from harassment caused by adversarial examples. Two defects exist in black-box iterative attacks that generate adversarial examples by incrementally adjusting the noise-adding direction for each step. On the one hand, existing iterative attacks add noises monotonically along the direction of gradient ascent, resulting in a lack of diversity and adaptability of the generated iterative trajectories. On the other hand, it is trivial to perform adversarial attack by adding excessive noises, but currently there is no refinement mechanism to squeeze redundant noises. In this work, we propose Curls \& Whey black-box attack to fix the above two defects. During Curls iteration, by combining gradient ascent and descent, we `curl' up iterative trajectories to integrate more diversity and transferability into adversarial examples. Curls iteration also alleviates the diminishing marginal effect in existing iterative attacks. The Whey optimization further squeezes the `whey' of noises by exploiting the robustness of adversarial perturbation. Extensive experiments on Imagenet and Tiny-Imagenet demonstrate that our approach achieves impressive decrease on noise magnitude in $\ell_{2}$ norm. Curls \& Whey attack also shows promising transferability against ensemble models as well as adversarially trained models. In addition, we extend our attack to the targeted misclassification, effectively reducing the difficulty of targeted attacks under black-box condition.
\end{abstract}

\section{Introduction}

The output of deep neural networks (DNNs) is highly sensitive to tiny perturbation on input images~\cite{Szegedy2013IntriguingPO, goodfellow2014explaining}. Among all methods that generate adversarial examples, iterative attacks \cite{kurakin2016adversarial, dong2017boosting, wu2018understanding} strike a better balance between attack effect and efficiency of adversarial example generation. However, there are two severe drawbacks in current mainstream black-box iterative attacks based on substitute model \cite{Papernot2017PracticalBA}. In the first place, decision boundaries between models in black-box scenario are far apart \cite{Liu2016DelvingIT}. Iterative trajectories have difficulties crossing decision boundary of target model with a small noise magnitude, because they are based on monotonic search along the gradient ascent direction of substitute model. This impairs adversarial examples' transferability\cite{Liu2016DelvingIT}. In the second place, although noise magnitude determines the performance of attack methods, adversarial examples generated by iterative attacks contain a certain amount of redundant noises that cannot be completely removed by simply increasing the iteration number. A post-iteration refinement mechanism is needed to squeeze out the `whey' of adversarial noises.

%iterative attacks generate adversarial examples by exploiting the inherent vulnerability of substitute model. But adding noises monotonically along the gradient ascent direction of substitute model may cause difficulties for adversarial examples to transfer to target model in black-box scenarios.

\begin{figure}[t]
  \centering
    \includegraphics[width=0.98\linewidth]{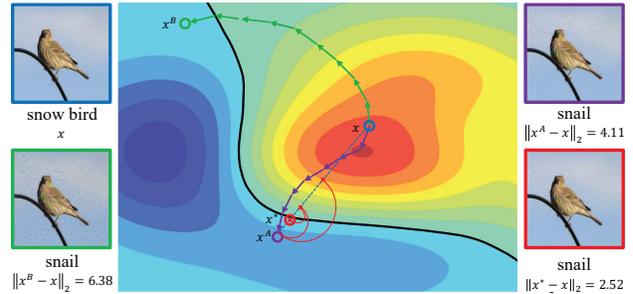}
  \caption{ Iterative trajectory of Curls iteration. Background is contour of cross entropy loss. The redder the color, the lower the loss. The consecutive black curve represents decision boundary between category `snow bird' and `snail'. Two polylines of green and purple represent iterative trajectories with simply gradient ascend and Curls iteration with binary search, respectively. The blue and red rings represent the original image $x$ and adversarial example found after binary search. Original image and three adversarial examples on both sides correspond to four rings with the same color as the image border. }
    \label{fig_RC_BB}
\end{figure}

In this paper, we propose Curls \& Whey black-box attack. During Curls iteration, we iterate along both the gradient ascent and descent directions of substitute model's loss function, as demonstrated by green and purple polylines in Fig. \ref{fig_RC_BB}. The dual-direction setting `curls' up the iterative trajectories and is hence more likely to cross target model's decision boundary at a closer distance, which effectively enhances the diversity as well as transferability of adversarial examples. Diminishing marginal effect caused by monotonically adding noises along the direction of gradient ascent is also weakened. Mechanisms to refine adversarial noises (red arc in Fig. \ref{fig_RC_BB}) and guide initial direction are included at the end and beginning of Curls iteration, respectively.

Whey optimization is applied to further squeeze the magnitude of noise by exploiting adversarial perturbation's robustness. We firstly divide adversarial perturbation into groups according to pixel value and attempt to filter out the noises of each group. Then we distill each pixel in adversarial example stochastically to squeeze out redundant noises little by little. Experiments on Imagenet~\cite{russakovsky2015imagenet} and Tiny-Imagenet \cite{brendel2018adversarial} verify that our method generates adversarial examples with higher transferability and smaller perturbation in $\ell_{2}$ norm under the same query limitation. We also systematically investigate the influence of each iterative parameter on the performance of the proposed method. In addition, our method shows strong transferability against ensemble models and adversarially trained models \cite{Tramr2017EnsembleAT}.

Targeted misclassification in black-box scenario has long been considered intractable~\cite{Liu2016DelvingIT}, for differences on decision boundaries and classification spaces between substitute and target model hampers adversarial examples' penetration from source class to target class. Most existing iterative attacks try to solve this problem by simply replacing gradient descent in untargeted misclassification with gradient ascent towards the target class~\cite{kurakin2016adversarial, dong2017boosting}. In this paper, by integrating interpolation to iterative process, we boost original image into the direction towards the target category and significantly decrease the difficulty of targeted misclassification.

We summarize our contributions as follows:

(1) We bring forward Curls iteration, a black-box attack method aiming at improving diversity of iterative trajectories and transferability of adversarial examples by combining both gradient ascent and gradient descent directions.

(2) We propose Whey optimization, the first noise-squeezing method exploiting robustness of perturbations.

(3) We expand our iterative method to targeted attacks and significantly improve attack effect of iterative methods under black-box scenario.

(4) We make our codes publicly available at \url{https://github.com/walegahaha/Curls-Whey}.

\section{Related Work} \label{section_related_work}

In black-box attack, attackers can only query target model and get the score of each category \cite{papernot2016limitations}. One practical solution exploits transferability between local substitute model and the target model, i.e., phenomenon that adversarial examples generated by one model can fool another \cite{Papernot2017PracticalBA}. Four existing attacks are introduced in the following.

\textbf{Fast Gradient Sign Method (FGSM).} As a classical one-step attack, FGSM \cite{goodfellow2014explaining} finds the noise's direction by calculating the gradient of cross-entropy loss $J(x,y_T)$:
\begin{equation}\label{FGSM}
  x^{\prime}=x+\varepsilon\cdot sign(\bigtriangledown J(x,y_T)).
\end{equation}

\textbf{Iterative FGSM (I-FGSM).} I-FGSM \cite{kurakin2016adversarial} splits uppper bound of noise $\varepsilon$ into several small step size $\alpha$ and adds noises step by step:
\begin{equation}\label{I-FGSM}
  \quad x_{t+1}^{\prime} = Clip_{x ,\varepsilon}\{x_t^{\prime}+\alpha\cdot sign(\bigtriangledown J(x_t^\prime,y_T))\}.
\end{equation}

I-FGSM possesses the highest attack effect among all current iterative attacks in white-box scenario. Its main drawback is the diminishing marginal effect of iterative steps. In other words, as the number of iterations $t$ increases and the step size $\alpha$ decreases, keeping adding the iteration step has little improvement on attack effect.

\textbf{Momentum Iterative FGSM (MI-FGSM).} MI-FGSM \cite{dong2017boosting} introduced a momentum term to make the adjustment of the noise-adding direction smoother, but the impact of diminishing marginal effect on iteration number still exists:
\begin{eqnarray}\label{MI-FGSM}
  m_{t+1} &=& \mu\cdot m_t+\frac{\bigtriangledown J(x_t^\prime,y_T))}{\ \ \| \bigtriangledown J(x_t^\prime,y_T))\|},  \\
  x_{t+1}^{\prime} &=& Clip_{x,\varepsilon}\{x_t^{\prime}+\alpha\cdot sign(g_{t+1})\}.
\end{eqnarray}

\textbf{Variance-Reduced Iterative FGSM (vr-IGSM).} Vr-IGSM \cite{wu2018understanding} replaces the gradient of the original image with an averaged gradient of original image with gaussian noises.
\begin{eqnarray}\label{vr-MI-FGSM}
  G_{t+1} &=& \frac{1}{m} \sum_{i=1}^{m} \bigtriangledown J(x_t+\xi_i),\quad \xi_i\sim\mathcal{N}(0, \sigma^2 I),  \\
  x_{t+1}^{\prime} &=& Clip_{x,\varepsilon}\{x_t^{\prime}+\alpha\cdot sign(G_{t+1})\}.
\end{eqnarray}

Gaussian noise eliminates local fluctuation in substitute model, thus improves the transferability.

A series of defence methods have been proposed to improve robustness of target models \cite{papernot2016distillation, li2017adversarial, meng2017magnet}. Among them, adversarial training \cite{Tramr2017EnsembleAT} and model ensemble are two most widely-used methods. Adversarial training vaccinates against adversarial examples by including them into the training set of target model, while model ensemble reduces specific error made by single model.

\section{Curls \& Whey Attack} \label{section_iterative_strategy}

%In this section, we propose a novel iterative attack consisting two phases of Roller Coaster iteration and ebb optimization. We first specify the formal expression of iterative adversarial attack. Then we show the insights for Roller Coaster iteration and ebb optimization. After that, we describe these two phases in detail. Finally, we expand our iterative method to targeted attacks under black-box scenario.

\subsection{Notation}
\label{notation}

An image classifier based on DNN can be represented as $N: X^{W\times H\times C}\rightarrow Y^K$, where $X$ represents the input space with dimension of $Width\times Height\times Channel$ and $Y$ represents the classification space with $K$ categories. A successful adversarial attack changes the original classification result of image classifier, i.e., the target model, after adding as little noise as possible to the original image \cite{wang2016theoretical}:
\begin{equation}\label{attack_definition}
  min  \ \| x^\prime-x\|_v, \quad s.t. \ N(x)\neq N(x^\prime)\ ,
\end{equation}
where $v$ refers to the norm used to measure the noise magnitude including $\ell_{1}$, $\ell_{2}$ and $\ell_\infty$ norm. In this paper we discuss noise magnitude in $\ell_{2}$ norm. Some existing works~\cite{xie2018improving, zhou2018transferable, dong2017boosting} compare the misclassification rate with a fixed $\ell_\infty$ norm, but we concentrate on the quality of adversarial noises generated by different attacks on one image. Here the black-box attack using substitute model \cite{Papernot2017PracticalBA} is used to solve the problem that the target model cannot be back propagated. The gradient information at step $t$ refers to the gradient value of the substitute model's loss function $J_{sub}$, i.e., cross-entropy loss, to adversarial example $x^\prime_t$.

\subsection{Diminishing Marginal Effect on Iteration Steps}
\label{subsection_marginal}

\begin{figure}[t]
  \centering
    \includegraphics[width=0.96\linewidth]{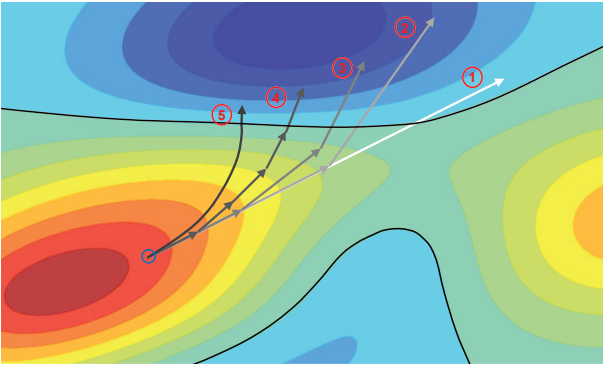}
  \caption{Diminishing marginal effect on iteration number $T$. The small blue ring at the bottom left represents the original image. Five polylines marked  $\textcircled{\tiny{1}}$ - $\textcircled{\tiny{5}}$ are iterative trajectories for $T=1, 2, 3, 5, \infty$ cross the decision boundary.}
    \label{fig_marginal}
\end{figure}

Iterative attacks perform well in white-box scenarios, where the transferability is guaranteed to be 100\% \cite{papernot2016transferability}. However, when attacking against a black-box target model, the drawbacks of iterative attacks gradually expose. First of all, discrepancy on decision boundary burdens transferability between substitute model and target model \cite{tramer2017space}. Iterative attacks always step toward the direction in which loss function of substitute model increases. But there is a huge gap on classification spaces between different models. Their gradient directions may be even orthogonal to each other \cite{Liu2016DelvingIT}. Therefore, simply searching for adversarial examples along the gradient ascent direction of the substitute model may no longer be suitable for black-box attacks.

What's more, diminishing marginal effect on the number of iterations exists. Now assume that in order to minimize the noise magnitude, the step size $\alpha$ of each step is inversely proportional to the total iteration numbers. In I-FGSM, when the number of iterations $T$ increases by 1, the marginal gain for the decrease in the noise magnitude is
\begin{equation}\label{equation_marginal}
  \sum_{t=1}^{T+1} \frac{1}{T+1} \cdot \bigtriangledown J_{sub}(x_t) - \sum_{t=1}^{T} \frac{1}{T} \cdot \bigtriangledown J_{sub}(x_t).
\end{equation}

In general, as $T$ increases and the single step size shortens, the iterative trajectory tends to be consistent and smooth and gradually converges, as shown in Fig. \ref{fig_marginal}. Considering that the number of queries to the target model in black-box attack is also limited, increasing the iteration number has little effect on adversarial noise reducing if the iteration number is already high.

\subsection{Curls Iteration}
\label{subsection_curls}

Iterative trajectories of current iterative attacks in black-box scenario are monotonic. First, monotonically employing gradient ascent along substitute model's loss function is more likely to bring iterative trajectories into local optimum of substitute model, rather than passing through the decision boundary of target model. Second, simply relying on transferability between substitute model and target model, but ignoring the feedback of target model after each query makes the iterative trajectories lack adaptability.

To `curl' up and diversify the iterative trajectory may be a more cost-effective solution \cite{sethi2018data}. Fig. \ref{fig_RC_BB} shows one possible distribution of target model loss function. In the case that loss function rises slowly along the direction of gradient ascend, like the green trajectory, it may be possible to find a shortcut across the decision boundary from a nearby starting point, as shown by the purple polyline in Fig. \ref{fig_RC_BB}. We abandon the monotonic search strategy base on gradient ascend to increase the diversity of iterative trajectories:

\begin{algorithm}[t]
\caption{Curls Iteration} %算法的名字
\hspace*{0.02in} {\bf Input:} %算法的输入， \hspace*{0.02in}用来控制位置，同时利用 \\ 进行换行
Target DNN $N(x)$, substitute model $Sub(x)$\\
\hspace*{0.42in} Original image $x$ and label $y$\\
\hspace*{0.42in} Initial noise magnitude limit $\varepsilon$\\
\hspace*{0.42in} Iteration step $T$ and variance of gaussian noise $s$ \\
\hspace*{0.42in} Step size $\alpha$ and binary search step $bs$\\
\hspace*{0.02in} {\bf Output:} %算法的结果输出
Adversarial example $x^\prime$
\begin{algorithmic}[1]

\State Initialize $\bar{R}$ and two starting points  \\
$\bar{R} = 0, x_0^A=x, x_0^B=x$
\State $downhill = True$ \textit{\small $\//\//$ Set the gradient descend flag to True}
\For{$t$ = 0 to $T$}
    \State $ \xi_t^A, \xi_t^B \sim\mathcal{N}(0, s^2 I) $
    \State Calculate gradient on substitute model
    \State $g_t^A = \bigtriangledown J_{sub}(x_t^A + \xi_t^A +  \alpha\cdot \bar{R}) $
    \State $g_t^B = \bigtriangledown J_{sub}(x_t^B + \xi_t^B +  \alpha\cdot \bar{R}) $
    \State $$ x_{t+1}^A = \left\{
    \begin{array}{rcl}
        Clip_{x,\varepsilon}\{x_t^A - \alpha\cdot g_t^A \} & &{downhill = True} \\
        Clip_{x,\varepsilon}\{x_t^A + \alpha\cdot g_t^A \} & &{downhill \neq True} \\
    \end{array} \right.$$

    $ x_{t+1}^B = Clip_{x,\varepsilon}\{x_t^B + \alpha\cdot g_t^B \} $

    \If{ $downhill = True \ and \ J(x_{t+1}^A) > J(x_{t}^A) $ }
        \State $downhill = False$
    \EndIf

    \If{ $ N(x_{t+1}^A)\neq N(x) \ or \ N(x_{t+1}^B)\neq N(x) $}
        \State update $\bar{R}$ by Eqn. \eqref{equation_RO}
    \EndIf
\EndFor
\If{ $ N(x_T^A)\neq N(x) \ or \ N(x_T^B)\neq N(x) $}
    $$x^\prime=\left\{
    \begin{array}{rcl}
        x_T^A & &{ \| x_T^A - x\|_2 < \| x_T^B - x\|_2 }\\
        x_T^B & &{else}\\
    \end{array} \right.$$
    \State refine $x^\prime$ by Eqn. \eqref{equation_BB}
\EndIf
\State \Return $x^\prime$
\end{algorithmic}
\label{alg:1}
\end{algorithm}

\begin{eqnarray}\label{equation_RC}
  x_0^\prime &=& x, \ x_1^\prime = Clip_{x ,\varepsilon}\{x_0^{\prime}-\alpha\cdot \bigtriangledown J_{sub}(x_0^\prime)\}, \\
  g_{t+1} &=& \left\{
  \begin{aligned}
  -\bigtriangledown J_{sub}(x_t^\prime) \quad  & J(x_t^\prime)<J(x_{t-1}^\prime),\\
   \bigtriangledown J_{sub}(x_t^\prime) \quad  & J(x_t^\prime)\geq J(x_{t-1}^\prime),\\
  \end{aligned}
  \right.  \\
  x_{t+1}^{\prime} &=& Clip_{x,\varepsilon}\{x_t^{\prime}+ \alpha\cdot g_{t+1} \},
\end{eqnarray}
where $J_{sub}(x_t^\prime)$ and $J(x_t^\prime)$ represent the cross entropy loss of adversarial example $x_t^\prime$ on the substitute model and the target model, respectively. First, update the original image for one step along the direction of gradient descent. When the cross entropy loss of current adversarial example on target model is lower than the previous step, usually the `valley floor', i.e., the local minimum of loss function has not yet been reached. Therefore, when the loss on the target model is still declining, continue to update along the direction of gradient descend, and vice versa. We regard this `first go down then go up' iterative method as Curls iteration.

On the basis of Curls, we introduce two heuristic strategies before and after each round of iteration. For an image, the closest adversarial examples are more likely to distribute in roughly the same direction in the feature space. Therefore, we record and update the average direction of all adversarial examples of one image, $\bar{R}$, and adds a vector pointing to this direction in the first step when calculating gradients for each round:
\begin{align}\label{equation_RO}
  & \bar{R}  = \frac{1}{K} \sum_{i=1}^{K} x^\prime , \quad s.t. \ N(x)\neq N(x^\prime)\ ,  \\
  & x_1^\prime  = Clip_{x,\varepsilon}\{x_0^{\prime}+\alpha\cdot \bigtriangledown J(x_0^\prime +  \alpha\cdot \bar{R})\}.
\end{align}

Since the iterative trajectory cannot be a straight line in the high-dimensional feature space, situation shown in the red arcs in Fig. \ref{fig_RC_BB} exists: there are adversarial examples with smaller $\ell_{2}$ distance between the adversarial example found and original image. We perform binary search between original image $x$ and adversarial example $x^\prime$ after each round to fully exploit the potential of this round:
\begin{align}
  & L = x, R = x^\prime , \\
  & BS(L, R) = \left\{
  \begin{aligned}\label{equation_BB}
    & BS(L, (L + R)/2) , \\
    & if \ N(x) \neq N((L + R)/2) , \\
    & BS((L + R)/2, R) , \\
    & if \ \quad N(x) = N((L + R)/2) . \\
  \end{aligned}
  \right.
\end{align}

In the actual implementation of Curls iteration, in order to prevent the oscillation of adversarial noise update, we do not directly determine the gradient symbol on account of target model's loss function, but divide each iterative round into two stages. In the first stage, carry out gradient descend to the original image. Once the cross entropy on target model is lower than the previous step, the second stage starts and carries out gradient ascend until the last step. At the same time, the normal iterative trajectory of direct gradient ascent is performed simultaneously. In addition, inspired by vr-IGSM~\cite{wu2018understanding}, we add gaussian noise to image in gradient calculation process to improve the transferability. Algorithm \ref{alg:1} details Curls iteration.

\subsection{Whey Optimization}
Usually an iterative attack ends as soon as it finds adversarial example or runs out of iteration number. However, adversarial examples generated may still contain redundant `whey' noises after iteration. Or the maximum extent to which noises can be reduced, while ensuring the adversarial example can still fool the target model \cite{athalye2017synthesizing}:
$$max(\parallel x^\prime-x\parallel_2-\parallel x^\circ-x\parallel_2), \quad s.t.\quad N(x^\prime)=N(x^\circ),$$
where $x$, $x^\prime$ and $x^\circ$ refers to original image, adversarial example found by now and the closest adversarial example to the original image, respectively.

Since binary search between $x$ and $x^\prime$ is already performed, adversarial examples with less redundant noises are more likely to exist in a linearly independent direction with respect to $x^\prime - x$. We propose Whey optimization to squeeze out the remaining `whey' of redundant noises in black-box attack. Whey optimization maintains a balance between noise-squeezing amplitude and the number of squeezes. Squeezing excessive noises at a time may return adversarial examples to the original category. Nevertheless, an incremental squeeze makes it impossible for optimization to complete within a limited number of queries. A compromise solution is to divide adversarial noises into groups first, then try to reduce noise magnitude group by group:
\begin{eqnarray}\label{equation_GO}
   z_0 &=& x^\prime-x , \\
   z_{t+1}^{whc} &=& z_{t}^{whc}/2, \quad  s.t. \ z_{t}^{whc} = L(V(z_0),t) ,
\end{eqnarray}
\begin{algorithm}[t]
\caption{Whey Optimization} %算法的名字
\hspace*{0.02in} {\bf Input:} %算法的输入， \hspace*{0.02in}用来控制位置，同时利用 \\ 进行换行
Target DNN $N(x)$ and adversarial example $x^\prime$\\
\hspace*{0.42in} Original image $x$ and label $y$\\
\hspace*{0.42in} Max attempt number for two squeeze steps, $T_1$, $T_2$\\
\hspace*{0.42in} Pixel value set of $x^\prime - x$, $P$\\
\hspace*{0.42in} Random number generator over $[0, 1]$, random() \\
\hspace*{0.02in} {\bf Output:} %算法的结果输出
Refined adversarial example $x^\ast$
\begin{algorithmic}[1]
\State $z=x^\prime-x$
\State $t_1=0$, $t_2=0$
\For{$p$ in $P$ and $t_1<T_1$} \textit{\small $\//\//$ Step 1: Squeeze in groups}
    \State Reduce the pixel value by half
    \State \quad$z\left[z=p\right]/=2$
    \If{$N(z) = y$}
        \State  Cancel the update of this step
    \EndIf
    \State $t_1=t_1+1$
\EndFor
\While{$t_2<T_2$}  \textit{\small $\//\//$ Step 2: Squeeze stochastically}
    \State Generate a random mask same shape as the image
    $$mask^{whc}=\left\{
    \begin{array}{rcl}
        0 & &{random() \leq0.01,}\\
        1 & &{else.}\\
    \end{array} \right.$$

    \State $z = z\cdot mask$  \textit{\small $\//\//$ Element-wise product}

    \If{$N(z) = y$}
        \State  Cancel the update of this step
    \EndIf
    \State $t_2=t_2+1$
\EndWhile
\State $x^\ast=z+x$
\State \Return $x^\ast$
\end{algorithmic}
\label{alg:2}
\end{algorithm}
where $z$ is the noise, $L(V,t)$ represents number with the $t^{th}$ largest absolute value in pixel value set $V$:
$$V(z)=\{v\mid v=z^{whc},\ w\in[0,W],h\in[0,H],c\in[0,C]\}$$
$W,H,C$ represents the width, height and channel of original image $x$, respectively. Whey optimization divides noise $z$ into several groups according to the pixel value, selects one group each time in descending order, reduces all pixel value in $z$ which equals to $L(V,t)$ by half and check whether the trimmed noises can still fool the target model.

After squeezing in groups, we perform more fine-grained squeeze. The last step of Whey optimization set the value of each pixel to 0 with probability of $\delta$:
\begin{eqnarray}\label{equation_SD}
    z_{t+1} &=& z_{t}\cdot mask_{t}, \\
    mask^{whc} &=& \left\{
        \begin{array}{rcl}
            0 & &{random()\leq\delta},\\
            1 & &{else},\\
        \end{array} \right.
\end{eqnarray}
where $mask$ is the same shape as $z$. Algorithm \ref{alg:2} gives the detail of Whey optimization.

\subsection{Targeted Attack}  \label{subsection_targeted_method}

Unlike untargeted attack, targeted attack requires not only the adversarial example be misclassified by the target model, but also it can be misclassified into the specified category. This is especially difficult in black-box attack because the decision boundaries between different models vary greatly, and the gradient direction are even orthogonal to each other \cite{Liu2016DelvingIT}. Even if the update of each step is changed from gradient ascend with respect to the original category $\bigtriangledown J_{sub}(x^\prime,y_{ori})$ to gradient descend with respect to the target category $- \bigtriangledown J_{sub}(x^\prime,y_{target})$ \cite{dong2017boosting}, an iterative trajectory from original image is almost impossible to reach the target category space, due to the difference in gradient values between target model and substitute model.

We abandon the `start from scratch' strategy and integrate interpolation to the iterative attack to get a better initial update direction. First, we collect a legitimate image $x_T$ that can be classified into the target category by the target model. Second, we use binary search to find an image $x_0^\prime$ between the original image $x$ and $x_T$, making sure that $x_0^\prime$ can also be classified into the target category. After that, we use $x_0^\prime$ to guide the first gradient ascent step starting from $x$:
\begin{eqnarray}\label{equation_targeted}
   x_0^\prime &=& (1-s) \cdot x + s \cdot x_T, \\
   x_1^\prime &=& Clip_{x ,\varepsilon}\{x - \alpha\cdot \bigtriangledown J(x_0^\prime)\}, \\
   x_{t+1}^\prime &=& Clip_{x ,\varepsilon}\{x_t^{\prime}-\alpha\cdot \bigtriangledown J(x_t^\prime)\}, t\geq1,
\end{eqnarray}
where $0<s<1$ indicates the interpolation coefficient determined by binary search. In this way, we boost original example into the direction towards the target category. After the first boosting step, we continue to apply Curls\&Whey attack as in untargeted attacks.

\section{Experiments} \label{section_experiments}

\subsection{Experiment Settings} \label{subsection_experiment_settings}

\begin{table*}[th]
  \caption{Median and average $\ell_{2}$ distance of adversarial perturbation crafted from pairwise attack between four models.}
  \label{table_tiny_untargeted}
  \small
  \centering
  \begin{tabular}{c|c|cc|cc|cc|cc}
    \toprule
    & & \multicolumn{2}{c|}{resnet18} &\multicolumn{2}{c|}{inceptionv3} &\multicolumn{2}{c|}{inception resnet v2} &\multicolumn{2}{c}{nasnet}  \\
    \midrule
    & attack methods & median & average & median & average & median & average & median & average  \\
    \midrule
                    & FGSM      &  \textit{0.1321} & \textit{0.8893}  &  4.3085          & 7.4580          & 3.6764          & 5.3257          & 3.4187          & 4.5589 \\
                    & I-FGSM    &  \textit{0.0800} & \textbf{\textit{0.0881}}&  1.9686   & 2.9287          & 2.4624          & 3.3192          & 2.1865          & 2.9644 \\
    resnet 18       & MI-FGSM   &  \textit{0.0866} & \textit{0.1029}  &  2.3220          & 3.4386          & 2.9526          & 3.9267          & 2.0174          & 2.9723 \\
                    & vr-IGSM   &  \textit{0.0941} & \textit{0.1120}  &  1.8737          & 2.8228          & 2.4803          & 3.4085          & 1.7991          & 2.7645 \\
                &\textbf{Curls} &  \textit{0.0731} & \textit{0.1182}  &  1.6443         & 2.4739          & 1.8507          & 2.6290          & 1.6773          & 2.4919 \\
       &\textbf{Curls\&Whey}&\textit{\textbf{0.0627}}& \textit{0.1040}& \textbf{1.1942}& \textbf{1.7387} & \textbf{1.4549} & \textbf{1.9450} & \textbf{1.3902} & \textbf{1.9696} \\
    \midrule
                    & FGSM      &  0.9944          & 3.6262           &  \textit{0.1521}          & \textit{1.9010}          & 2.6171          & 4.9078          & 2.8729          & 4.5217 \\
                    & I-FGSM    &  0.6699          & 1.8883           &  \textbf{\textit{0.1132}} & \textbf{\textit{0.1518}} & 1.3415          & 1.9095          & 1.3774          & 2.1675 \\
inception v3        & MI-FGSM   &  0.8124          & 2.2895           &  \textit{0.1283}          & \textit{0.1989}          & 1.6248          & 2.4642          & 1.6800          & 2.7336 \\
                    & vr-IGSM   &  0.6072          & 1.7973           &  \textit{0.1297}          & \textit{0.1834}          & 1.3214          & 2.0991          & 1.3569          & 2.3010 \\
                    &\textbf{Curls}&  0.5760       & 1.6781           &  \textit{0.1243}          & \textit{0.2194}          & 1.1163          & 1.8997          & 1.2335          & 2.1067 \\
                    &\textbf{Curls\&Whey}&  \textbf{0.5140} & \textbf{1.4941}  &  \textit{0.1252}          & \textit{0.9200}          & \textbf{0.9058} & \textbf{1.7913} & \textbf{0.9398} & \textbf{1.9315} \\
    \midrule
                    & FGSM      &  1.6729          & 5.0270           &  4.2482          & 6.6191          & \textit{0.2855}          & \textit{4.5974}          & 4.1107          & 5.5487 \\
                    & I-FGSM    &  0.7019          & 2.3966           &  1.3314          & 2.3834          & \textbf{\textit{0.1293}} & \textit{0.3814}          & 1.3761          & 2.3732 \\
inception resnet v2 & MI-FGSM   &  0.8561          & 2.8611           &  1.6342          & 3.0884          & \textit{0.1602}          & \textit{0.5419}          & 1.6594          & 3.0469 \\
                    & vr-IGSM   &  0.6463          & 2.4453           &  1.3166          & 2.6256          & \textit{0.1640}          & \textit{0.5197}          & 1.3292          & 2.6710 \\
                    &\textbf{Curls}&  0.6040       & 2.0220           &  1.1325          & 1.9407          & \textit{0.1501}          & \textit{0.3450}          & 1.0978          & 1.9644 \\
                    &\textbf{Curls\&Whey}&\textbf{0.5227}& \textbf{1.2404}& \textbf{0.8431}& \textbf{1.3437}& \textit{0.1485}         & \textbf{\textit{0.3199}} & \textbf{0.8483} & \textbf{1.4403} \\
    \midrule
                    & FGSM      &  3.7356          & 6.0550           &  3.5277          & 7.2388          & 3.4829          & 7.1657          & \textit{0.2008}          & \textit{6.3891} \\
                    & I-FGSM    &  1.5575          & 4.1401           &  1.5926          & 4.3745          & 1.4180          & 4.2968          & \textbf{\textit{0.1173}} & \textit{1.8225} \\
nasnet              & MI-FGSM   &  0.9518          & 3.0544           &  1.8850          & 3.9685          & 1.6458          & 3.7643          & \textit{0.1317}          & \textit{0.3632} \\
                    & vr-IGSM   &  0.5659          & 2.4410           &  1.5006          & \textbf{3.2440} & 1.3066          & \textbf{3.1112} & \textit{0.1371}          & \textbf{\textit{0.3197}} \\
                    &\textbf{Curls}&  0.5821       & 2.1520           &  1.2719          & 3.9490          & 1.2048          & 4.1637          & \textit{0.1360}          & \textit{2.7491} \\
                    &\textbf{Curls\&Whey}&\textbf{0.5543}& \textbf{1.8582}&\textbf{1.0003}& 3.6760         & \textbf{0.9599} & 3.6069          & \textit{0.1354}          & \textit{2.5653} \\
    \bottomrule
  \end{tabular}
\end{table*}

All our experiments are performed on Tiny-Imagenet used in NIPS 2018 Adversarial Vision Challenge \cite{brendel2018adversarial} and Imagenet \cite{russakovsky2015imagenet}, with image shape of $64\times64\times3$ and $224\times224\times3$, respectively. Imagenet contains 1000 image categories. We picked 10000 images from its validation set that can be correctly classified by all target models, 10 images for each category. As for Tiny-Imagenet with 200 image categories, we choose 2000 images, 10 images for each category. 8 neural network models with different structures are compared: resnet-18 \cite{he2016deep}, resnet-101, inception v3 \cite{szegedy2016rethinking}, inception-resnet v2~\cite{szegedy2017inception}, nasnet \cite{Zoph2017LearningTA}, densenet-161 \cite{Huang2017DenselyCC}, vgg19-bn \cite{simonyan2014very}, senet-154 \cite{hu2017}.

We implement our black-box iterative attack on Foolbox \cite{rauber2017foolbox} framework. In order to accurately measure the attack effect of each method, a large loop for determining $\varepsilon$ is added outside the iterative process. For evaluation criterion, we choose the median and average size of adversarial perturbation transferred from substitute model to target model, as applied in NIPS 2018 Adversarial Vision Challenge \cite{brendel2018adversarial}:
\begin{eqnarray}\label{equation_metric}
    mid(Sub, N) &=& median( \{ d(x, x^\ast) \mid x\in \textbf{X} \}), \\
    avg(Sub, N) &=& \frac{1}{N} \sum_{i=1}^{N}(\{ d(x, x^\ast) \mid x\in \textbf{X} \}), \\
    d(x, x^\ast) &=&  \ \| x - x^\ast \|_2 ,
\end{eqnarray}
where $sub$ and $N$ represent substitute model and target model, respectively. $x$ is an original image in the test set $\textbf{X}$. $x^\ast$ is the adversarial example found that is closest to $x$. $d(x, x^\ast)$ returns the $\ell_{2}$ distance between $x$ and $x^\ast$. A smaller $\ell_{2}$ distance indicates a stronger attack effect and higher the transferability of generated adversarial examples.

\subsection{Black-box Attack on Multiple Models} \label{subsection_big_black_exp}

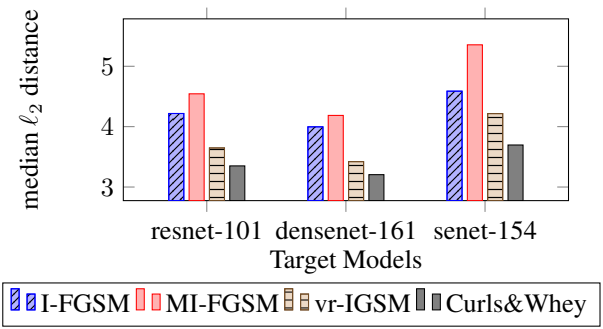
\begin{figure}[t]
  \centering
    \begin{tikzpicture}
    \begin{axis}[
        height=4cm,
        width=7.5cm,
        xlabel=Target Models,
    	ylabel=median $\ell_{2}$ distance,
        xtick=data,
        %tickwidth= 0pt,
    	enlargelimits=0.3,
    	legend style={at={(0.4,-0.45)},anchor=north,legend columns=-1},
        %nodes near coords,
        enlarge y limits  = 0.2,
        symbolic x coords={resnet-101, densenet-161, senet-154},
        ybar,
        bar width=0.2cm]

        \addplot +[postaction={ pattern=north east lines }]
        	coordinates {(resnet-101, 4.217926747)  (densenet-161, 3.996998214) (senet-154, 4.587466017)};
        \addplot
        	coordinates {(resnet-101, 4.543804397)  (densenet-161, 4.18611783) (senet-154, 5.355194505)};
        \addplot +[postaction={ pattern=horizontal lines}]
        	coordinates {(resnet-101, 3.647490967)  (densenet-161, 3.420426102) (senet-154, 4.216081423)};
        \addplot
        	coordinates {(resnet-101, 3.350034131)  (densenet-161, 3.204943899) (senet-154, 3.696183364)};
        \legend{I-FGSM, MI-FGSM, vr-IGSM, Curls\&Whey}
    \end{axis}
    \end{tikzpicture}
  \caption{Median $\ell_{2}$ distance comparison of adversarial noises generated using vgg19-bn as substitute model on Imagenet.}
  \label{fig_imagenet_vgg19}
\end{figure}

We report the median and average adversarial perturbation on Tiny-Imagenet in Table \ref{table_tiny_untargeted}. In this $4 \times 4$ matrix, each element represents the result of substitute model of this row against the target model of this column over the entire 2000 images. Elements on diagonal are results of white-box attacks (marked in italics). Fig. \ref{fig_imagenet_vgg19} shows median perturbation on three target models when using vgg19-bn as substitute model. More experiments on Imagenet can be found in supplemental material. For each pair of substitute and target model, we compare our methods (Curls\&Whey as well as Curls only) with FGSM~\cite{goodfellow2014explaining} and three other iterative attacks, I-FGSM \cite{kurakin2016adversarial}, MI-FGSM \cite{dong2017boosting} and vr-IGSM \cite{wu2018understanding}. Since $\ell_{2}$ norm is used to measure noise magnitude, we no longer use sign function to update adversarial examples. For the fairness of comparison, the number of queries to the target model is basically equal for the iterative attacks. Table \ref{table_exp_set} reports parameters related to query number, including iterative round number $T_0$, iteration step $T$, binary search step $bs$, max attemp number for two squeeze steps in Whey optimization $T_1$ and $T_2$. The total query number for our method is $T_0\times(T+bs)\times 2+T_1+T_2$, and $T_0\times T$ for other iterative methods. The initial noise magnitude $\varepsilon$ and stepsize $\alpha$ are 0.3 and $1/2T$, respectively. For variance of gaussian noise in vr-IGSM and our method, we set $s=1$.

\begin{table}[t]
  \caption{Parameter set for experiments on two datasets.}
  \label{table_exp_set}
  \small
  \centering
  \begin{tabular}{c|c|c|c|c|c|c|c}
    \toprule
    & & $T_0$ & $T$ & $bs$ & $T_1$ & $T_2$ & Total  \\
    \midrule
    Tiny-          & Others    &  20   & 10  & -- & -- & -- & 200 \\
    Imagenet       & Ours      &  10   & 4   & 2  & 40 & 40 & 200 \\
    \midrule
    Imagenet       & Others    &  24   & 24  & -- & -- & -- & 576 \\
                   & Ours      &  14   & 7   & 3  &200 &100 & 580 \\
    \bottomrule
  \end{tabular}
\end{table}

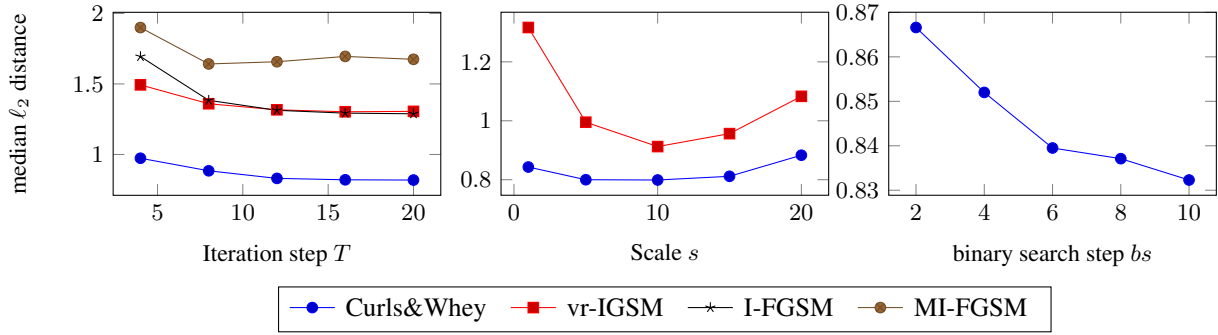
\begin{figure*}[th]
  \begin{center}
  \begin{tikzpicture}

  \begin{groupplot}[
      group style={group size=3 by 1,ylabels at=edge left, horizontal sep=0.8cm},
      ylabel style={text height=0.01\textwidth,inner ysep=0pt},
      height=0.23\linewidth,width=0.34\linewidth,/tikz/font=\small
    ]

    \nextgroupplot [xlabel={Iteration step $T$}, ylabel={median $\ell_{2}$ distance}]
    \addplot  coordinates {
        (4      ,0.9732)
        (8      ,0.8842)
        (12     ,0.8304)
        (16     ,0.8204)
        (20     ,0.8185)
    };\label{plots:RCIM+Ebb}
    \addplot  coordinates {
        (4      ,1.4931)
        (8      ,1.3595)
        (12     ,1.3165)
        (16     ,1.3026)
        (20     ,1.3055)
    };\label{plots:vr-IGSM}
    \addplot  coordinates {
        (4      ,1.8995)
        (8      ,1.6415)
        (12     ,1.6572)
        (16     ,1.695)
        (20     ,1.6744)
    };\label{plots:MI-FGSM}
    \addplot  coordinates {
        (4      ,1.6941)
        (8      ,1.3839)
        (12     ,1.3129)
        (16     ,1.2929)
        (20     ,1.2883)
    };\label{plots:I-FGSM}
    \coordinate (top) at (rel axis cs:0,1);% coordinate at top of the first plot
    \nextgroupplot [xlabel={Scale $s$}]
    \addplot  coordinates {
        (1      ,0.843106911)
        (5      ,0.8)
        (10     ,0.7991)
        (15     ,0.8116)
        (20     ,0.8832)
    };\label{plots:RCIM+Ebb_2}
    \addplot  coordinates {
        (1     ,1.3166)
        (5     ,0.9952)
        (10    ,0.9125)
        (15    ,0.9564)
        (20    ,1.083)
    };\label{plots:vr-IGSM_2}
    \nextgroupplot [xlabel={binary search step $bs$}]
    \addplot  coordinates {
        (2      ,0.8666)
        (4      ,0.852)
        (6     ,0.8395)
        (8     ,0.8371)
        (10     ,0.8323)
    };\label{plots:RCIM+Ebb_3}
    \coordinate (bot) at (rel axis cs:1,0);% coordinate at bottom of the last plot
  \end{groupplot}
  % legend
  \path (top|-current bounding box.south)--
        coordinate(legendpos)
        (bot|-current bounding box.south);
  \matrix[
      matrix of nodes,
      anchor=north,
      draw,
      inner sep=0.2em,
    ]at([yshift=-1ex]legendpos)
    {
      \ref{plots:RCIM+Ebb}& Curls\&Whey&[5pt]
      \ref{plots:vr-IGSM} & vr-IGSM&[5pt]
      \ref{plots:I-FGSM}& I-FGSM&[5pt]
      \ref{plots:MI-FGSM}& MI-FGSM&[5pt]\\};
\end{tikzpicture}

\end{center}
\caption{Median noise magnitude under different iteration steps (left), $T=(4,8,12,16,20)$ , gaussian noise variance (middle), $s=(1, 5, 10, 15, 20)$ and binary search step (right), $bs=(2, 4, 6, 8, 10)$.}
    \label{fig_parameters}
\end{figure*}

\begin{figure*}[t]
  \centering
    \includegraphics[width=0.96\linewidth]{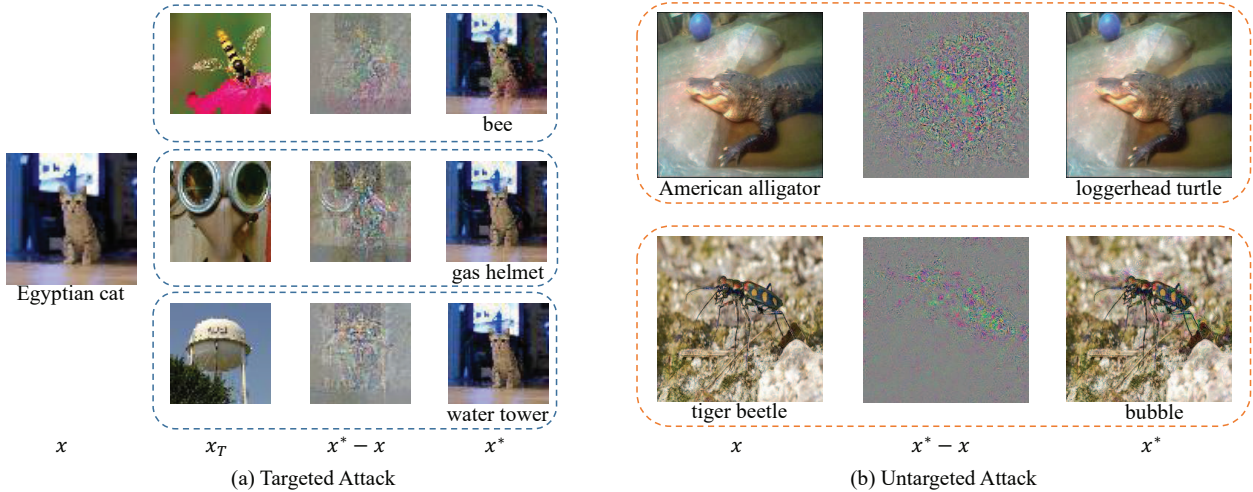}
  \caption{Adversarial examples generated by Curls \& Whey attack. Targeted attack results on Tiny-Imagenet are shown on subplot (a). Original image $x$, image of target category $x_T$, noise $x^\ast-x$ and adversarial example $x^\ast$ are listed from left to right. Untargeted results on Imagenet are shown on subplot (b). Classification result on target model are shown at the bottom.}
    \label{fig_example}
\end{figure*}

It can be seen from Table \ref{table_tiny_untargeted} that Curls\&Whey achieves smaller median noise magnitude in $\ell_{2}$ norm than all other methods, and smaller average magnitude than most other methods, on black-box attacks, i.e., off-diagonal elements. With the diversification of iterative trajectories and squeeze of redundant noises, noises are reduced by 20\%-30\%, in some cases even 40\%, over most model combinations. Curls iteration alone also outperforms existing methods in almost all black-box attacks. Due to gaussian noises in gradient-calculating process, noise magnitude of our methods are slightly higher than I-FGSM in white-box attacks, where transferability is no need to be considered. However, white-box noise of our method is still smaller than that of vr-IGSM, which validates the effectiveness of Whey optimization. Fig. \ref{fig_example} shows adversarial examples crafted on two datasets. Curls \& Whey achieves targeted and untargeted misclassification with nearly imperceptible noises.

\subsection{Ablation Study} \label{subsection_ablation}

Here we investigate influence of iteration step $T$, binary search step $bs$ and variance of gaussian noise $s$ to black-box attack effect. We use inception-resnet v2 and inception v3 as substitute and target model, respectively. Results on Tiny-Imagenet under different $T$, $s$ and $bs$ is shown in Fig. \ref{fig_parameters}. As discussed in Section \ref{section_iterative_strategy}, although $T$ is negatively correlated with noise magnitude, diminishing marginal effect exists. The noise drop of $T=20$ relative to $T=16$ is obviously not as great as the drop of $T=8$ relative to $T=4$. Our method does not simply increase the iteration number, but improve the diversity of iterative trajectories. Therefore, Curls\&Whey is able to find adversarial examples with smaller $\ell_{2}$ norm with equal queries, and use part of the query to refine adversarial noises.

Variance $s$ is related to the transferability between substitute and target model. The higher the $s$, the greater the likelihood that adversarial example may transfer from one model to another highly different model. However, as the variance of gaussian noise increases, the proportion of original image in gradient calculation process will gradually decrease, resulting in decline in transferability. Therefore, a local minimum appears in the results on different $s$. As can be seen from Fig. \ref{fig_parameters}, when using inception-resnet v2 to attack inception v3, the local optimal value of $s$ is around 10.

As for binary search step, a larger $bs$ means more binary search between the adversarial example and original image. As an auxiliary process in Curls iteration, a relatively small $bs$ is sufficient to reduce the noises.
\begin{table}[t]
  \caption{Incremental comparison on each part of Curls\&Whey.}
  \label{table_ablation_part}
  \centering
  \begin{tabular}{c|c|c|c|c}
    \toprule
    & Curls & +BS & +Whey(1) & +Whey(2)   \\
    \midrule
    median          & 1.3138 &  1.1111   & 0.9354  & \textbf{0.8431}  \\
    \midrule
    average        & 2.3154  &  1.9039   & 1.4723  & \textbf{1.3437}  \\
    \bottomrule
  \end{tabular}
\end{table}

To verify the effectiveness of each part of our attack method, we conduct ablation experiment on Curls\&Whey. As can be seen from Table \ref{table_ablation_part}, whether it is Curls iteration, binary search (BS), or two steps in Whey optimization, each component can effectively reduce the noise magnitude.

\subsection{Targeted Attack Results} \label{subsection_targeted}

\begin{figure}[t]
    \begin{tikzpicture}
    \begin{axis}[
        height=3.5cm,
        width=7.5cm,
    	ylabel=median $\ell_{2}$,
        ymin=0,
        ymax=88,
        xtick=data,
        %tickwidth= 0pt,
    	enlargelimits=0.3,
        %nodes near coords,
        %enlarge x limits  = 1,
        enlarge y limits  = 0,
        symbolic x coords={inceptionv3, inc-resnet v2, nasnet},
        ybar,
        bar width=0.15cm]

        \addplot +[postaction={ pattern=north east lines }, yellow]
        	coordinates {(inceptionv3, 81.2345)  (inc-resnet v2, 80.9559) (nasnet, 81.9787)};
        \addplot
        	coordinates {(inceptionv3, 80.9999)  (inc-resnet v2, 80.6864) (nasnet, 81.6331)};
        \addplot +[postaction={ pattern=horizontal lines}]
        	coordinates {(inceptionv3, 80.9087)  (inc-resnet v2, 80.6796) (nasnet, 81.6413)};
        \addplot +[postaction={ pattern=north west lines}]
        	coordinates {(inceptionv3, 39.8188)  (inc-resnet v2, 39.9544) (nasnet, 40.0107)};
        \addplot +[postaction={ pattern=north east lines}]
        	coordinates {(inceptionv3, 31.2757)  (inc-resnet v2, 31.5086) (nasnet, 31.4495)};
        \addplot
        	coordinates {(inceptionv3, 24.8444)  (inc-resnet v2, 24.0634) (nasnet, 24.2808)};
        \addplot +[postaction={ pattern=horizontal lines}]
        	coordinates {(inceptionv3, 9.3365)   (inc-resnet v2, 9.1087)  (nasnet, 9.2421)};

    \end{axis}
    \end{tikzpicture}

    \begin{tikzpicture}
    \begin{axis}[
        height=3.5cm,
        width=7.5cm,
    	ylabel=median $\ell_{2}$,
        xtick=data,
        ymin=0,
        ymax=88,
        %tickwidth= 0pt,
    	enlargelimits=0.3,
        enlarge y limits  = 0,
    	legend style={at={(0.5,-0.4)},anchor=north,legend columns=3},
        %nodes near coords,
        %enlarge x limits  = 1,
        symbolic x coords={resnet18, inc-resnet v2, nasnet},
        ybar,
        bar width=0.15cm]

        \addplot +[postaction={ pattern=north east lines }, yellow]
        	coordinates {(resnet18, 81.7994)  (inc-resnet v2, 80.6272) (nasnet, 81.8072)};
        \addplot
        	coordinates {(resnet18, 81.7709)  (inc-resnet v2, 80.6065) (nasnet, 81.2065)};
        \addplot +[postaction={ pattern=horizontal lines}]
        	coordinates {(resnet18, 81.5944)  (inc-resnet v2, 80.5404) (nasnet, 81.6332)};
        \addplot +[postaction={ pattern=north west lines}]
        	coordinates {(resnet18, 40.0754)  (inc-resnet v2, 39.9544) (nasnet, 40.0107)};
        \addplot +[postaction={ pattern=north east lines}]
        	coordinates {(resnet18, 31.7736)  (inc-resnet v2, 31.5086) (nasnet, 31.4495)};
        \addplot
        	coordinates {(resnet18, 27.1537)  (inc-resnet v2, 24.0634) (nasnet, 24.2808)};
        \addplot +[postaction={ pattern=horizontal lines}]
        	coordinates {(resnet18, 9.3832)   (inc-resnet v2, 7.0783)  (nasnet, 7.9913)};
        \legend{I-FGSM, MI-FGSM, vr-IGSM, pointwise, boundary attack, interpolation, Curls\&Whey}
    \end{axis}
    \end{tikzpicture}

  \caption{Median $\ell_{2}$ distance comparison of targeted adversarial noises generated using resnet18 (up) and inceptionv3 (down) as substitute model on Tiny-Imagenet.}
  \label{fig_targeted}
\end{figure}
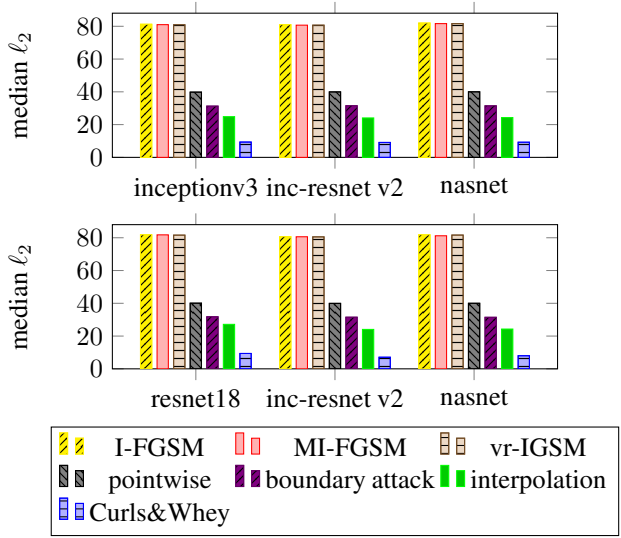

\begin{table}[t]
  \caption{Median and average $\ell_{2}$ distance of adversarial perturbation against adversarially trained models and ensemble model.}
  \label{table_defence}
  \small
  \centering
  \begin{tabular}{c|ccc}
    \toprule
    target model & attack methods & median & average   \\
    \midrule
                    & FGSM      &  6.5812          & 9.1681           \\
                    & I-FGSM    &  2.8839          & 3.76             \\
    inceptionv3(adv)& MI-FGSM   &  3.8039          & 4.6529           \\
                    & vr-IGSM   &  3.2752          & 4.1449           \\
                    &Curls\&Whey&  \textbf{2.0633} & \textbf{2.6349}  \\
    \midrule
                    & FGSM      &  4.7029          & 6.2954           \\
                    & I-FGSM    &  3.3195          & 3.9606           \\
    inc-resnet v2(adv)& MI-FGSM &  3.9919          & 4.9481           \\
                    & vr-IGSM   &  3.3829          & 4.2706           \\
                    &Curls\&Whey&  \textbf{2.2852} & \textbf{2.7884}  \\
    \midrule
                    & FGSM      &  4.5826          & 5.9755           \\
    inceptionv3+    & I-FGSM    &  2.7742          & 3.595           \\
    inc-resnet v2+  & MI-FGSM   &  3.5819          & 4.5227           \\
    nasnet          & vr-IGSM   &  3.0785          & 4.0499           \\
                    &Curls\&Whey&  \textbf{2.0321} & \textbf{2.6187}  \\
    \bottomrule
  \end{tabular}
\end{table}

In experiments on targeted attack, we assign 5 different target categories for each image and calculate the $\ell_{2}$ distance between original image and adversarial examples of each target category. As discussed in Section \ref{subsection_targeted_method}, we select one image from the test set that can be classified into target category for interpolation. We choose resnet18 and inceptionv3 as our substitute model and three other models as target models. As can be seen from the Fig. \ref{fig_targeted}, three existing iterative attacks have difficulties achieving targeted misclassification with small $\ell_{2}$ distance in black-box scenario. Compared to three decision-based attacks, boundary attack \cite{brendel2017decision}, pointwise attack and vanilla interpolation \cite{rauber2017foolbox}, noise magnitude of our method is also significantly reduced. This confirms the effectiveness of integrating interpolation method into Curls \& Whey attack.

\subsection{Attack on Defence and Ensemble Models} \label{subsection_defence}
Adversarial training \cite{Tramr2017EnsembleAT} and model ensemble are two widely used defend methods. In Table \ref{table_defence}, we use resnet18 as substitute models to attack two adversarially trained models (inceptionv3 and inception-resnet v2) and ensemble model consisting of three models. Although defence methods increase the difficulty of adversarial attack compared with Table~\ref{table_tiny_untargeted}, the noise magnitude of adversarial examples built by Curls \& Whey is still much lower than other attacks.

\section{Conclusion} \label{section_conclusion}
We propose Curls \& Whey, a new black-box attack containing Curls iteration and Whey optimization, to diversify the iterative trajectory and squeeze the adversarial noises respectively. In addition, we integrate interpolation to iterative attack to reduce the difficulty of targeted attacks in black-box scenario significantly. Experimental results on Tiny-Imagenet and ImageNet demonstrate that compared to existing iterative attacks, Curls \& Whey generates adversarial examples with smaller $\ell_{2}$ distance and stronger transferability against a variety of target models.

{\small
\bibliographystyle{ieee}
\bibliography{egbib}
}

\clearpage
\renewcommand\thesection{\Alph{section}}
\setcounter{section}{0}

\begin{center}
  \section*{Supplemental Material}
\end{center}

In this supplemental material, we show additional experimental results and more adversarial examples generated by Curls \& Whey attack, including tables of untargeted black-box attack on Imagenet and targeted black-box attack on Tiny-Imagenet. Adversarial examples generated on two datasets are listed behind.

\section{Untargeted Attack}

\begin{figure*}[t]
  \centering
    \includegraphics[width=0.98\linewidth]{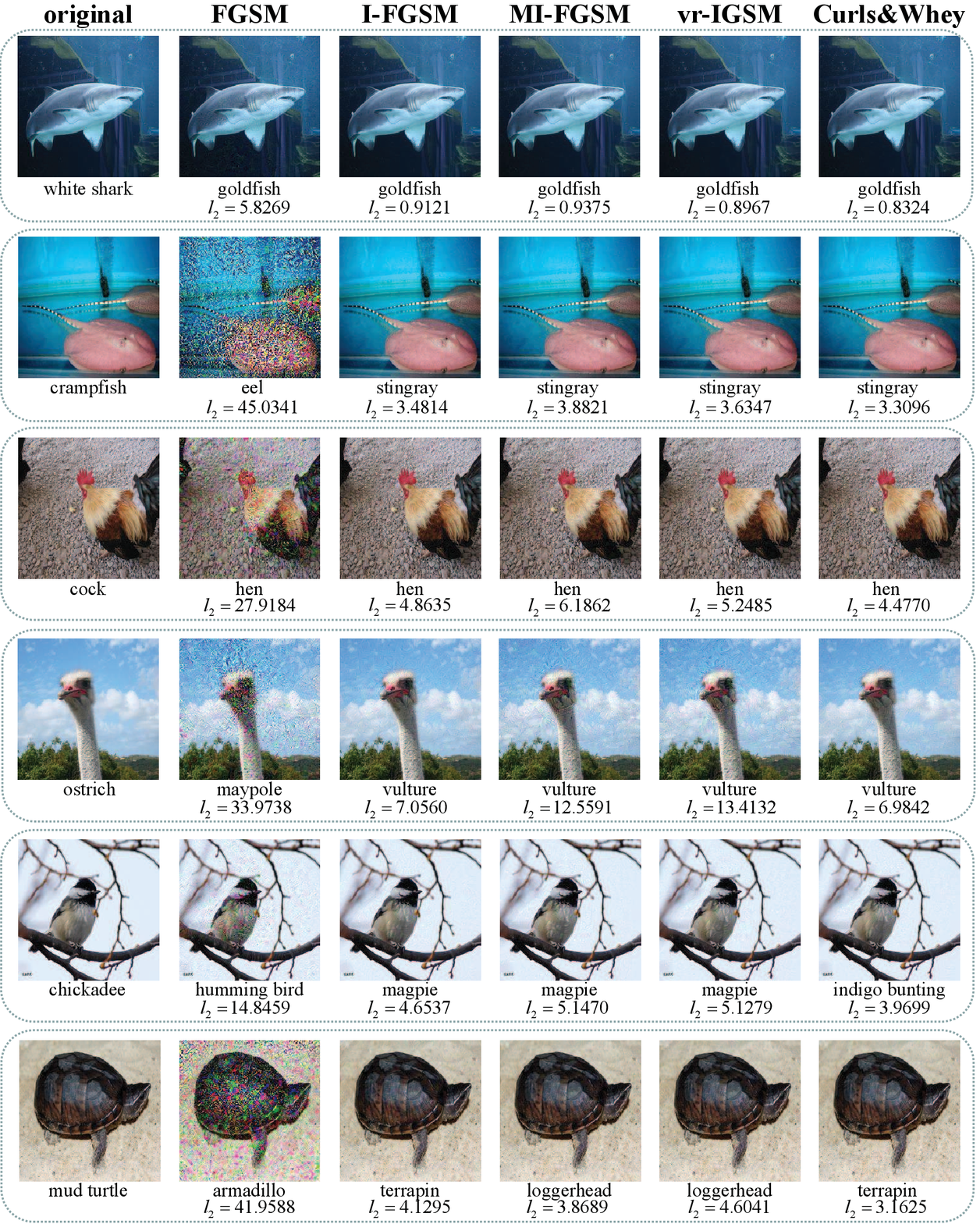}
  \caption{Six groups of untargeted attack examples on Imagenet. Columns from left to right show original image and the adversarial examples generated by five different methods. The misclassification category and $\ell_2$ norm of the noise are below each image.}
    \label{fig_first}
\end{figure*}

In Table \ref{table_imagenet_untargeted}, we report median and average $\ell_2$ distance of adversarial perturbations crafted on Imagenet dataset. Four DNN models with different structures are compared: resnet-101, densenet-161, vgg19-bn and senet-154. In this $4\times4$ matrix, each element represents the result of substitute model of this row against the target model of this column over the entire $1000\times10$ images collected from validate set of Imagenet, 10 images for each category. We compare our Curls \& Whey attack with four other attack methods, FGSM, I-FGSM, MI-FGSM and vr-IGSM. As can be seen, Curls \& Whey achieves smaller median noise magnitude in $\ell_2$ norm than other methods on black-box attacks (off-diagonal elements). Because of the gaussian noises introduced, our method as well as vr-IGSM perform not so well on white-box results (diagonal elements), where transferability is guaranteed to be 100\%. Fig. \ref{fig_first} illustrates several adversarial examples crafted by different attack methods. The leftmost images in Fig. \ref{fig_first} are original images. Noise magnitude in $\ell_2$ norm is placed below each adversarial example.

\begin{table*}[t]
  \caption{Median and average $\ell_2$ distance of adversarial perturbation for untargeted attack on Imagenet.}
  \label{table_imagenet_untargeted}
  \small
  \centering
  \resizebox{!}{0.523\columnwidth}{
  \begin{tabular}{c|c|cc|cc|cc|cc}
    \toprule
    & & \multicolumn{2}{c|}{resnet-101} &\multicolumn{2}{c|}{densenet-161} &\multicolumn{2}{c|}{vgg19-bn} &\multicolumn{2}{c}{senet-154}  \\
    \midrule
    & attack methods & median & average & median & average & median & average & median & average  \\
    \midrule
                    & FGSM      &  \textit{0.3167}          & \textit{3.9776}           &  6.7431          & 16.5051         & 6.0534          & 15.1296         & 7.9076          & 15.6174 \\
                    & I-FGSM    &  \textit{\textbf{0.2045}} & \textit{0.4065}           &  2.0577          & 3.8353          & 1.9820          & 4.0847          & 3.5190          & 6.5371 \\
    resnet-101       & MI-FGSM   & \textit{0.2305}          & \textit{\textbf{0.2864}}  &  2.2312          & 4.3944          & 2.2569          & 5.1220          & 4.1914          & 8.1522 \\
                    & vr-IGSM   &  \textit{0.2390}          & \textit{0.2940}           &  1.9114          & 3.7779          & 1.8940          & 4.3436          & 3.3970          & 7.0107 \\
                    &Curls\&Whey&  \textit{0.2295}          & \textit{0.5198}           &  \textbf{1.8655} & \textbf{3.5871} & \textbf{1.7285} & \textbf{3.5187} & \textbf{2.9872} & \textbf{5.3745} \\
    \midrule
                    & FGSM      &  6.5304          & 15.3936          &  \textit{0.3070}          & \textit{4.2659}          & 5.3533          & 13.0053         & 6.8372          & 14.5869 \\
                    & I-FGSM    &  1.8307          & 3.8534           &  \textit{\textbf{0.2173}} & \textit{0.5338}          & 1.7311          & 3.6356          & 2.8061          & 5.2298 \\
    densenet-161     & MI-FGSM   &  1.9436          & 4.2498           &  \textit{0.2258}          & \textit{0.2051}          & 1.9164          & 4.4277          & 3.2574          & 6.3578 \\
                    & vr-IGSM   &  1.8994          & 3.7641           &  \textit{0.2576}          & \textit{\textbf{0.1834}} & 1.6656          & 3.8322          & 2.7925          & 5.6133 \\
                    &Curls\&Whey&  \textbf{1.7041} & \textbf{3.3246}  &  \textit{0.2494}          & \textit{0.7397}          & \textbf{1.5771} & \textbf{3.1351} & \textbf{2.4977} & \textbf{4.4188} \\
    \midrule
                    & FGSM      &  9.9305          & 19.8893          &  8.7631          & 16.3457         & \textit{0.1819}          & \textit{2.2736}          & 11.2227         & 23.8974 \\
                    & I-FGSM    &  4.2179          & 8.7935           &  3.9970          & 7.5216          & \textit{\textbf{0.1406}} & \textit{0.8352}          & 4.5875          & 8.7350 \\
    vgg19-bn        & MI-FGSM   &  4.5438          & 9.9437           &  4.1861          & 8.3386          & \textit{0.1468}          & \textit{0.2462}          & 5.3552          & 10.2055 \\
                    & vr-IGSM   &  3.6475          & 8.3765           &  3.4204          & 6.9270          & \textit{0.1537}         & \textit{\textbf{0.2357}} & 4.2161          & 8.7974 \\
                    &Curls\&Whey&  \textbf{3.3500} & \textbf{6.9225}  &  \textbf{3.2049} & \textbf{6.3321} & \textit{0.1511}          & \textit{0.8173}          & \textbf{3.6962} & \textbf{7.1415} \\
    \midrule
                    & FGSM      &  8.3359          & 15.4190          &  8.3936          & 15.1964         & 7.9624          & 14.7991         & \textit{0.6791}          & \textit{5.9169} \\
                    & I-FGSM    &  4.2529          & 7.9353           &  4.1996          & 7.4578          & 2.4439          & 4.8991          & \textit{0.3478}          & \textit{0.9178} \\
    senet-154        & MI-FGSM   &  4.5414          & 9.9268           &  4.5520          & 9.6595          & 2.9568          & 6.6679          & \textit{0.4465}         & \textit{0.4386} \\
                    & vr-IGSM   &  3.4674          & 8.4754           &  3.5301          & 8.3745          & 2.5631          & 5.9603          & \textit{0.3226}          & \textit{0.7623} \\
                    &Curls\&Whey&  \textbf{3.0064} & \textbf{5.8348}  &  \textbf{3.0913} & \textbf{5.5426} & \textbf{1.9326} & \textbf{3.6826} & \textit{\textbf{0.2665}} & \textit{\textbf{0.4206}} \\
    \bottomrule
  \end{tabular}
  }
\end{table*}

\section{Targeted Attack}

\begin{figure*}[t]
  \centering
    \includegraphics[width=0.98\linewidth]{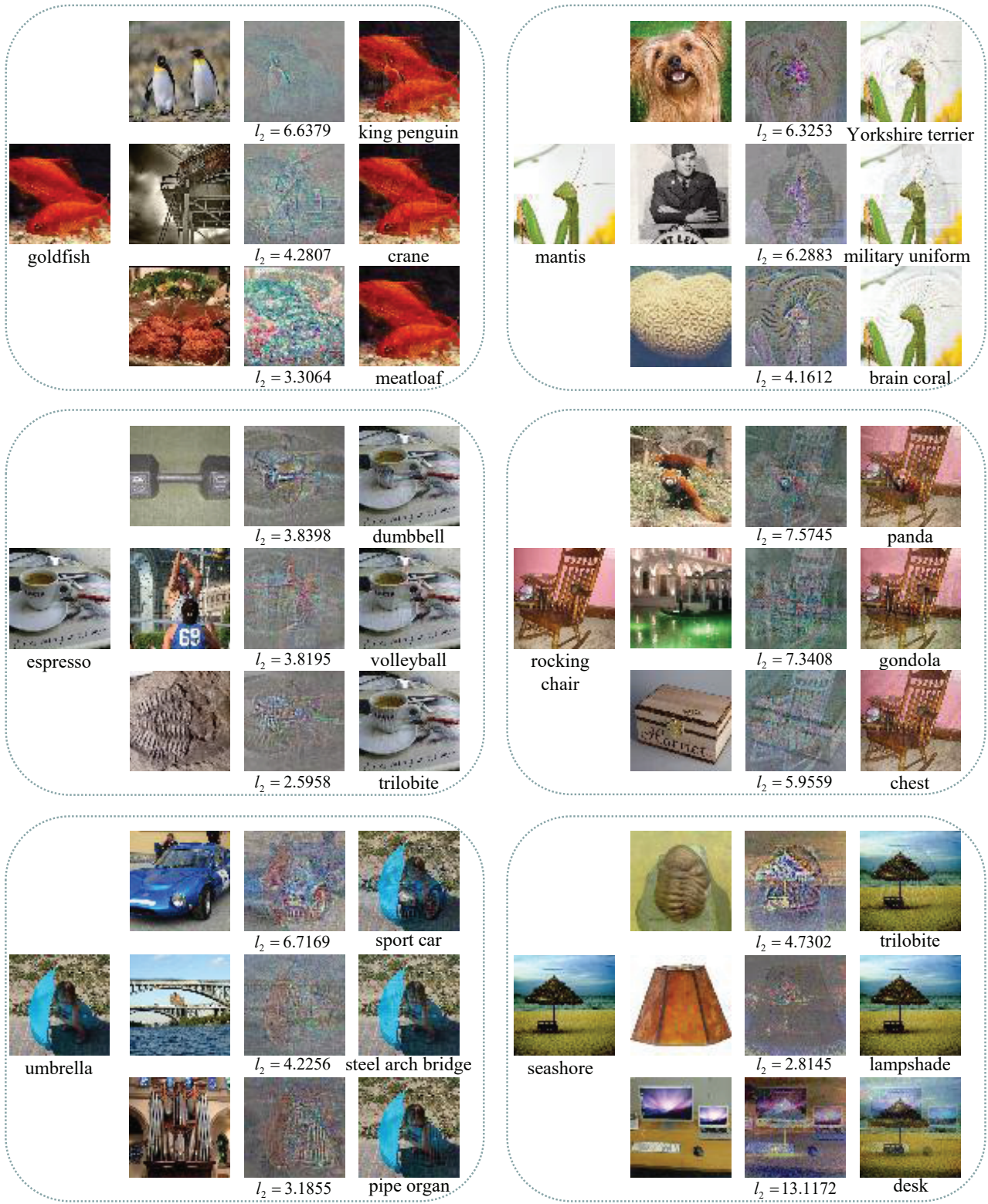}
  \caption{Six groups of targeted attack examples on Tiny-Imagenet. Original image, images of target category, noises and adversarial examples are listed from left to right in each group. By adding three different noises, each original image is misclassified into three other categories.}
    \label{fig_second}
\end{figure*}

In Table \ref{table_tiny_targeted} ,we provide median and average adversarial perturbation on $200\times10$ images collected from Tiny-imagenet dataset, 10 images for each category. Four DNN models are compared: resnet-18, inception V3, inception-resnet V2 and nasnet. As we can see, the performance of Curls \& Whey is far beyond other methods. The result of simply using iterative attacks like I-FGSM, MI-FGSM and vr MI-FGSM are all around 80, which means these methods seldom successfully achieve targeted misclassification with small $\ell_2$ distance. Three decision based attacks, boundary attack, pointwise attack and vanilla interpolation are also compared. These methods do not rely on substitute model, but collect a legitimate image that can be classified into the target category by the target model first and then search between original image and this image. Our method significantly reduces the noise magnitude of targeted attack in black-box scenario. Several groups of targeted adversarial examples are shown in Fig. \ref{fig_second}, where original image, image of target category, noise and targeted adversarial example are listed from left to right in each group.

\begin{table*}[t]
  \caption{Median and average $\ell_2$ distance of adversarial perturbation for targeted attack on Tiny-Imagenet.}
  \label{table_tiny_targeted}
  \small
  \centering
  \resizebox{!}{0.75\columnwidth}{
  \begin{tabular}{c|c|cc|cc|cc|cc}
  \toprule
    & & \multicolumn{2}{c|}{resnet-18} &\multicolumn{2}{c|}{inception V3} &\multicolumn{2}{c|}{inc-resnet V2} &\multicolumn{2}{c}{nasnet}  \\

    \midrule
                    & FGSM          & \textit{81.2735} & \textit{72.7756}          & 82.5241          & 80.9782          & 82.5322          & 81.0193         & 82.5626         & 80.9734  \\
                    & I-FGSM        & \textit{1.5398}  & \textit{2.9277}           & 81.2345          & 70.4633          & 80.9559          & 70.9525         & 81.9787         & 76.3114  \\
                    & MI-FGSM       & \textit{5.4267}  & \textit{34.7935}          & 80.9999          & 68.6765          & 80.6864          & 68.8399         & 81.6331         & 73.6505  \\
    resnet-18       & vr-IGSM & \textit{\textbf{0.3751}}  & \textit{\textbf{0.4328}}  & 80.9087     & 68.7406          & 80.6796          & 67.6213         & 81.6413         & 73.5731  \\
                    & Interpolation & \textit{27.1537}   & \textit{28.0997}          & 24.8444          & 25.2685          & 24.0634          & 24.9918         & 24.2808         & 24.9455  \\
                    & Pointwise     & \textit{40.0754}   & \textit{40.8887}    & 39.8188    & 40.4638     & 39.9544          & 40.6741         & 40.0107         & 40.6636  \\
                    & Boundary      & \textit{31.7736}   & \textit{32.5285}         & 31.2757      & 31.8612          & 31.5086          & 32.0049         & 31.4495         & 32.0101  \\
                    & Curls\&Whey   & \textit{2.9242}   & \textit{3.5819}          & \textbf{9.3365}  & \textbf{9.8224}  & \textbf{9.1087}  & \textbf{9.6767} & \textbf{9.2421} & \textbf{9.8868}\\
    \midrule
                    & FGSM        & 82.5945  & 81.4726          & \textit{82.5049}          & \textit{80.1433}          & 82.4776          & 80.8414         & 82.6294         & 81.7538  \\
                    & I-FGSM      & 81.7994  & 76.8168          & \textit{\textbf{0.3668}}  & \textit{\textbf{0.4005}}  & 80.6272          & 66.7824         & 81.8072         & 75.4738  \\
                    & MI-FGSM     & 81.7709  & 75.286           & \textit{0.6941}           & \textit{0.787}            & 80.6065          & 65.7128         & 81.2065         & 71.1222  \\
    inception V3    & vr-IGSM     & 81.5944  & 73.4011          & \textit{0.7074}           & \textit{0.7944}          & 80.5404          & 64.9861         & 81.6332         & 71.7732  \\
                    & Interpolation & 27.1537 & 28.0997         & \textit{24.8444}          & \textit{25.2685}          & 24.0634          & 24.9918         & 24.2808         & 24.9455  \\
                    & Pointwise     & 40.0754 & 40.8887         & \textit{39.8188}          & \textit{40.4638}          & 39.9544          & 40.6741         & 40.0107         & 40.6636  \\
                    & Boundary      & 31.7736  & 32.5285        & \textit{31.2757}          & \textit{31.8612}          & 31.5086          & 32.0049         & 31.4495         & 32.0101  \\
                    & Curls\&Whey   & \textbf{9.3832} & \textbf{10.2564} & \textit{1.2996}  & \textit{1.9423}    & \textbf{7.0783}  & \textbf{7.8715} & \textbf{7.9913} & \textbf{8.7716}\\
    \midrule
                    & FGSM          & 82.5945 & 81.1598          & 82.4787          & 80.9113          & \textit{82.4778}          & \textit{80.6907}         & 82.5397         & 81.4183  \\
                    & I-FGSM        & 81.8539 & 76.5748          & 80.5121          & 66.4706          & \textit{\textbf{0.5139}}  & \textit{\textbf{1.8753}} & 81.7709         & 74.8515  \\
                    & MI-FGSM       & 81.7926 & 75.2021          & 80.6945          & 66.7647          & \textit{1.5027}           & \textit{2.0268}          & 81.4114         & 71.1565  \\
    inc-resnet V2   & vr-IGSM    & 81.5944 & 74.2181          & 80.6342          & 66.0386          & \textit{1.5101}           & \textit{2.0193}          & 82.3142         & 75.3981  \\
                    & Interpolation & 27.1537 & 28.0997          & 24.8444          & 25.2685          & \textit{24.0634}          & \textit{24.9918}        & 24.2808         & 24.9455  \\
                    & Pointwise     & 40.0754 & 40.8887          & 39.8188          & 40.4638          & \textit{39.9544}         & \textit{40.6741}         & 40.0107         & 40.6636  \\
                    & Boundary      & 31.7736 & 32.5285          & 31.2757          & 31.8612          & \textit{31.5086}          & \textit{32.0049}         & 31.4495         & 32.0101  \\
                    & Curls\&Whey   & \textbf{9.4622} & \textbf{10.0514} & \textbf{7.7449}  & \textbf{8.3226}  & \textit{2.6911}           & \textit{3.0239}          & \textbf{7.5282} & \textbf{8.2552}\\
    \midrule
                    & FGSM          & 82.5626 & 81.3046  & 82.5626          & 81.4469          & 82.4840          & 81.0555         & \textit{82.4787}         & \textit{80.6151}  \\
                    & I-FGSM        & 81.6910 & 75.8341  & 79.8133          & 64.4298          & 79.9032          & 65.0252         & \textit{\textbf{0.3363}} & \textit{\textbf{0.3741}}\\
                    & MI-FGSM       & 81.6249 & 74.6014  & 80.5495          & 67.4740          & 79.8082          & 63.0784         & \textit{0.6785}          & \textit{0.7839}   \\
    nasnet          & vr-IGSM       & 81.6166 & 74.2587  & 80.5220          & 66.5961          & 79.7085          & 62.8439         & \textit{0.7582}          & \textit{0.8575}   \\
                    & Interpolation & 27.1537 & 28.0997  & 24.8444          & 25.2685          & 24.0634          & 24.9918         & \textit{24.2808}         & \textit{24.9455}  \\
                    & Pointwise     & 40.0754 & 40.8887  & 39.8188          & 40.4638          & 39.9544          & 40.6741         & \textit{40.0107}         & \textit{40.6636}  \\
                    & Boundary      & 31.7736          & 32.5285          & 31.2757          & 31.8612          & 31.5086          & 32.0049         & \textit{31.4495}         & \textit{32.0101}  \\
                    & Curls\&Whey   & \textbf{11.1441} & \textbf{11.867}  & \textbf{8.25630} & \textbf{8.70140} & \textbf{6.9883}  & \textbf{7.6393} & \textit{1.3578}          & \textit{1.9226}   \\

    \bottomrule
  \end{tabular}
  }
\end{table*}

\end{document}